\documentclass[preprint]{aastex631}
\usepackage{color}
\usepackage{amsmath}
\usepackage{verbatim}
\usepackage[section]{placeins}
\usepackage{xspace}

\newcommand{\autoscan}{\texttt{autoScan}}
\newcommand{\diffimg}{\texttt{DiffImg}}

\newcommand{\model}[1]{\textsf{#1}\xspace}

\begin{document}
\title{Transformer-Based Neural Network for Transient Detection without Image Subtraction}
\shorttitle{no-Diff R/B}

\author[0000-0002-1873-8973]{A. Inada}
\affiliation{Department of Physics and Astronomy, University of Pennsylvania, 209 South 33rd Street, Philadelphia, PA, 19104, USA}
\author[0000-0003-2764-7093]{M. Sako}
\affiliation{Department of Physics and Astronomy, University of Pennsylvania, 209 South 33rd Street, Philadelphia, PA, 19104, USA}
\author[0000-0002-5947-2454]{T. Acero-Cuellar}
\affiliation{Department of Physics and Astronomy, University of Delaware, 217 Sharp Lab, Newark, DE 19716, USA}
\author[0000-0003-1953-8727]{F. Bianco}
\affiliation{Department of Physics and Astronomy, University of Delaware, 217 Sharp Lab, Newark, DE 19716, USA}
\affiliation{University of Delaware, Joseph R. Biden, Jr. School of Public Policy and Administration, 184 Academy Street, Newark, DE 19716, USA}
\affiliation{University of Delaware, Data Science Institute, Newark, DE 19716, USA}
\affiliation{Vera C. Rubin Observatory, Tucson, AZ 85719, USA}

\correspondingauthor{A. Inada}
\email{inadas@sas.upenn.edu}
\shortauthors{The Roman Supernova Project Infrastructure Team}



\begin{abstract}
We introduce a transformer-based neural network for the accurate classification of real and bogus transient detections in astronomical images. This network advances beyond the conventional convolutional neural network (CNN) methods, widely used in image processing tasks, by adopting an architecture better suited for detailed pixel-by-pixel comparison. The architecture enables efficient analysis of search and template images only, thus removing the necessity for computationally-expensive difference imaging, while maintaining high performance. Our primary evaluation was conducted using the \autoscan\ dataset from the Dark Energy Survey (DES), where the network achieved a classification accuracy of 97.4\% and diminishing performance utility for difference image as the size of the training set grew. Further experiments with DES data confirmed that the network can operate at a similar level even when the input images are not centered on the supernova candidate. These findings highlight the network's effectiveness in enhancing both accuracy and efficiency of supernova detection in large-scale astronomical surveys.
\end{abstract}


\section{Introduction}

Identification of transient astronomical events such as supernovae (SNe) has played a crucial role in modern astronomy. In particular, type Ia SNe have confirmed the accelerating expansion of the universe \citep{perlmutter99, riess98} and ushered in an era of large-scale surveys designed to probe the expansion history of the universe such as Dark Energy Survey \citep[DES;][]{darkenergy2005}. The expanded opportunity set brought forth by improvements in telescopes, detectors, and computing power has increased the number of confirmed supernovae from less than 700 in 1988 to over 36,000 in 2017 \citep{asiago, openSN}.

Due to the rarity of SNe, however, the dramatic increase in the volume of available data has strained experts' ability to sift through the data and visually inspect the candidates to distinguish the real from ``bogus'' artifacts. A major challenge in efficiently detecting SNe lies in the difficulty of comparing corresponding sections of the sky across multiple epochs and isolating small variations in brightness. Since the vast majority of these variations are not scientifically important signals, there is a significant need to reduce the number of viable candidates. The standard approach to generating candidates is known as Difference Image Analysis (DIA). DIA involves a sequence of compute-intensive operations to perform image subtraction between $template$, a representative high quality image of the sky, and $search$, an image from a different epoch. Template images are created by co-adding high-quality aligned exposures \citep{alard, bramich2008}. The template image is then aligned to the search image and degraded to match the point spread function (PSF) of the search image. The final step subtracts the template from the search to only leave behind transients when applied correctly. While the methodology has continually improved since the original paper, DIA still suffers from a large number of artifacts that results from misalignment, spatial variations in the PSF, image defects, and differences in photometric scaling between the two images among others. Importantly, the calculation of the kernel to match PSF is a highly time-consuming step that limits the throughput of the difference image \citep{hu2022}.

The rise of machine learning in the field of astronomy offers an attractive alternative to improve the efficiency of real-bogus classification and significant progress has been made on this front in the past decade to cut down the number of candidates eligible for human scanning. This shift has lightened the load of manual inspection, making machine learning a vital component in processing the vast number of alerts, which for prominent surveys like Zwicky Transient Facility at Palomar, can be on the order of $10^6$ per night \citep{brink2013, mahabal2019}. Methods such as random forests, support vector machines, and deep neural networks have been employed to refine the selection process, leading to the development of classifiers with high accuracy over time \citep{bloom2012, goldstein2015, wright2015, morii2016, duev2019}. Within the context of deep neural networks, convolutional neural networks (CNNs) have emerged as prominent architectures, primarily due to their translation-equivariant property that is conducive to learning complex functions that take images as input.

While most of these works make use of difference image as part of their inputs, another recent work has shown that CNN can correctly classify supernova without difference image input at a modest accuracy loss of 5.0 percentage points (91.1\% vs 96.1\%) on the \autoscan\ dataset \citep{acero-cuellar2023}. The strength of this performance highlights the fact that difference image does not contain new information and real-bogus classification should only require search and template images in principle. Our work builds on this insight to explore an attention-based neural network that facilitates more efficient pixel-by-pixel comparison between search and template images to improve classification accuracy.

We will introduce the datasets used to train and evaluate the network in Section~\ref{sec:datasets}. Section~\ref{sec:network} discusses the neural network architecture and Section~\ref{sec:results} presents the results. Finally, we conclude in Section~\ref{sec:conclusions}. The networks used for this study are available on Github at \url{https://github.com/reiwanol/nodiff/}.

\newpage
\section{Datasets}
\label{sec:datasets}

\subsection{autoScan}

Our primary evaluation dataset is sourced from the \autoscan\ training set from the Dark Energy Survey, as detailed in the study by \cite{goldstein2015}. This dataset is invaluable particularly for benchmarking real-bogus classifiers in neural networks, owing to a substantial volume of realistic ``fake'' training data generated by \diffimg\ \citep{diffimg}. These images encapsulate a variety of astrophysical and instrumental contaminants such as defective pixels, cosmic rays, CCD edge effects, and alignment discrepancies, mirroring the complexities encountered in authentic astronomical data \citep{goldstein2015}.

In the realm of deep learning, the extensive dataset provided by DES is a valuable asset. Neural networks benefit significantly from large and diverse datasets when learning complex classification functions. The abundance of data in this dataset, featuring 898,963 visually inspected search-template-difference image samples, is vital for the effective training of these networks. The vast collection of data was amassed during the first operational year (Y1) of the DES-SN from August 2013 to February 2014. Images, with its 51 by 51-pixel resolution corresponding to about 14 arcseconds per side, offers a representative sample of the observing conditions.


In preparing the training data, we examined different approaches to pre-process the dataset. Specifically, 0 to 1 linear scaling, Gaussian scaling, and raw fits input approaches were studied. We observed raw fits input performed the best among three approaches, except for relatively small dataset of sample size below 25{,}000. The performance metrics mentioned all refer to networks trained on raw inputs unless otherwise specified.

Regarding the composition of the training dataset, a balanced approach was adopted. To avoid the introduction of bias, the dataset was reduced to include the same number of positive (real) and negative (bogus) samples, with each category comprising 400,384 training samples in the largest set. This equal distribution ensures that the neural network's training would not be skewed towards any particular classification, due to inherent bias in unbalanced training set. We split the dataset in 9:1 ratio between training and test sets in all of our training runs.

\subsection{no-Diff}
While the \autoscan\ dataset offers a strong benchmark to validate a real-bogus classifier, the search and template images indirectly use information from difference imaging pipeline as they are centered on sources detected by difference images. To avoid leveraging such information, we created a custom dataset from DES data that relies solely on sources extracted from search images. A total of 124,961 images captured between August 2013 and February 2018 in $griz$ bands were processed to create 1,763,176 search-template pairs. 

\begin{figure}[htbp]
    \centering
    \includegraphics[width=0.65\textwidth]{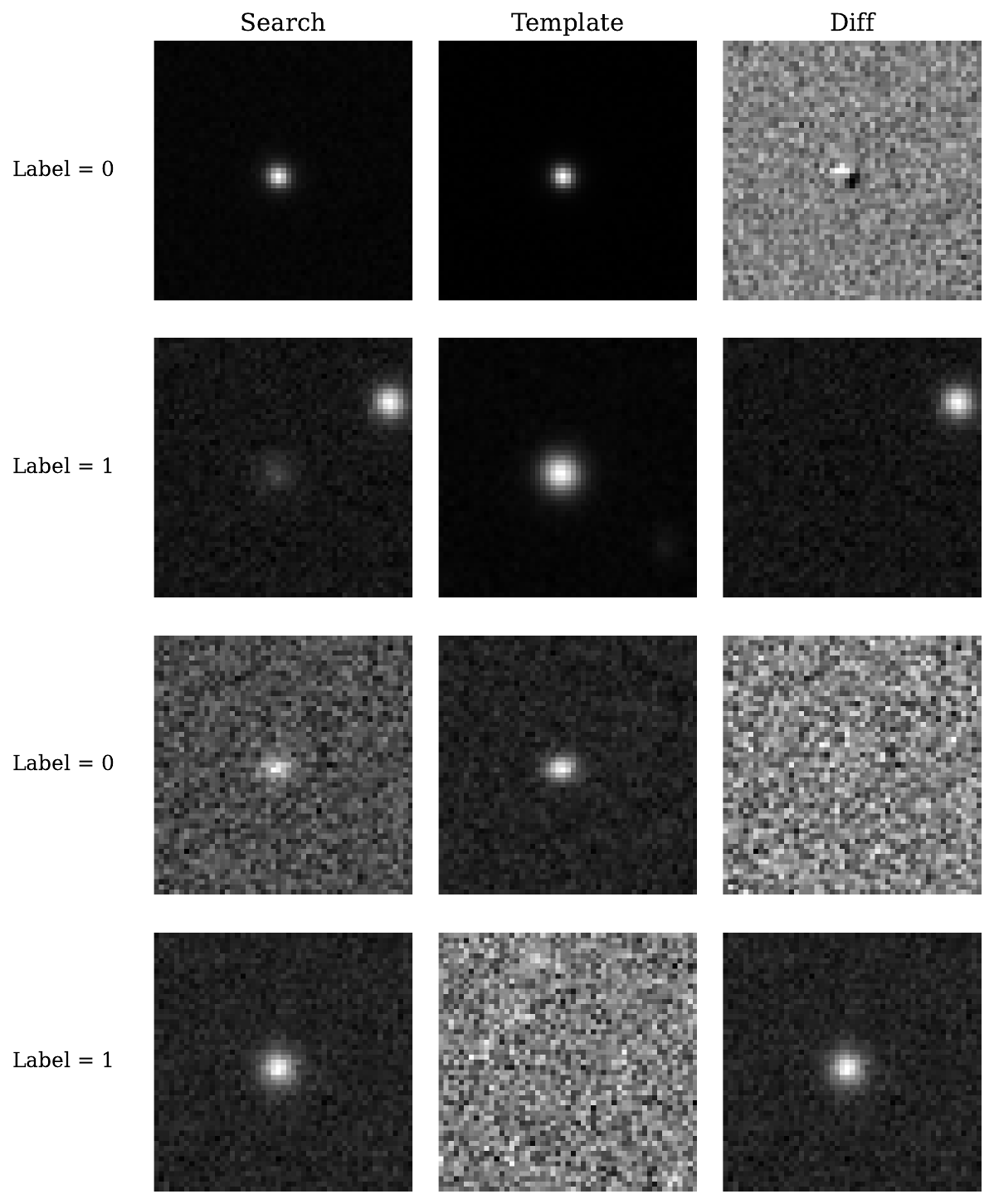}
    \caption{Stamps from no-Diff dataset. Positive (label = 1) and negative (label = 0) examples are shown. From the top row to the bottom row we show bad subtraction negative label, off the center positive label, clean subtraction negative label and cleanly subtracted positive label, respectively.}
    \label{fig:stamps}
\end{figure}

To build our custom no-Diff DES dataset, we created stamps from DES images containing injected fakes \citep{diffimg}. We extracted sources from our search images using Source Extractor at $\mathrm{SNR} > 5$. The corresponding stamps of dimension 51 by 51 pixels centered on detected source were then cut out from both search and template images. Search-template pairs containing fakes within the cut-out boundaries were labeled positive, and the rest were labeled negative. To make an appropriate comparison with the \autoscan\ dataset and focus on realistically retrievable fakes, only the stamps containing fake with available \autoscan\ scores and $\mathrm{SNR}$ greater than 3.5 from the difference image were kept for positive pairs. Up to 15 pairs per class were extracted from each of the images. Several stamps from the dataset are shown in Figure~\ref{fig:stamps}. The data is publicly available on NERSC\footnote{\url{https://portal.nersc.gov/cfs/dessn/nodiff/all.tar.gz}}.

Similar to the \autoscan\ dataset, we used raw inputs to train our neural network and split the dataset into training and test subset in a 9:1 ratio. Since the dataset is meaningfully larger than \autoscan\ and contains more negative pairs than positive pairs, we first balanced the dataset, randomly selected 20\% of the remaining dataset, and then performed the train-test split. This corresponds to dataset size of roughly 267,000 for each run.

To prevent data leakage between the training and test sets, we enforced a group-based split, ensuring that all stamps from a given exposure and CCD were assigned to either the training set or the test set, but not both.

\section{Network Architecture}
\label{sec:network}

Our network architecture draws inspiration from the success of attention mechanisms in image processing tasks. End-to-end attention-based models like Vision Transformer \citep[ViT;][]{dosovitskiy2020image} and DETR \citep{carion2020detr} have extended the applicability of attention mechanisms, originally popularized in natural language processing, to the domain of image processing. In particular, ViT has demonstrated superior performance compared to traditional convolutional neural networks (CNNs) on large datasets and its principles have been adapted for transient detection in astronomical imaging \citep{chen2023transientvit}.

By applying a modified attention mechanism to a search-template pair, our network introduces a more effective pixel-by-pixel comparison scheme suitable for real-bogus classification in the absence of a difference image. Specifically, the network localizes the attention mechanism by introducing a trainable parameter that influences the attention weights based on distances between pixels.

\begin{figure}[htbp]
    \centering
    \includegraphics[width=0.9\textwidth]{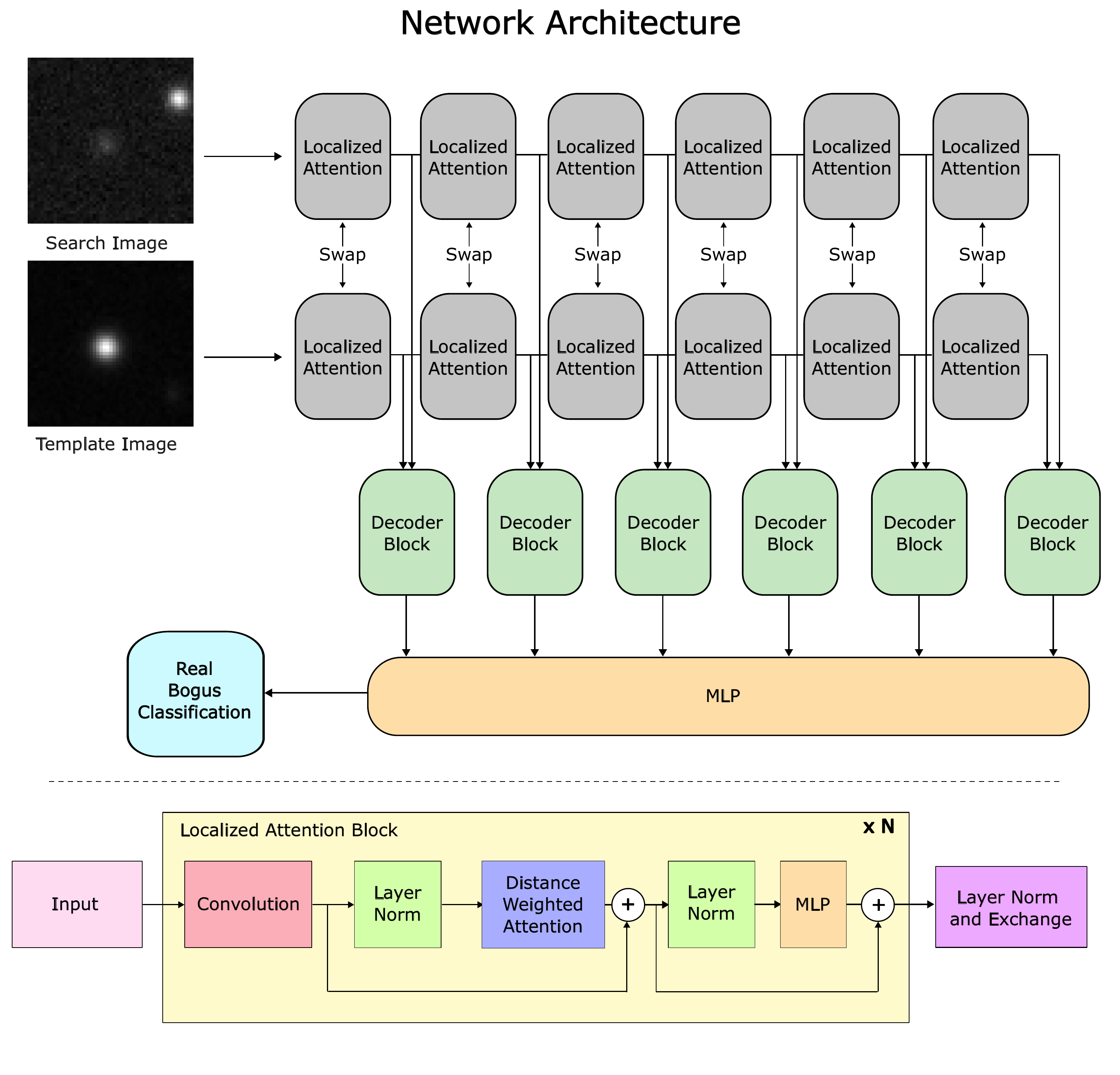}
    \caption{Network architecture of real-bogus classifier used in this work. (\textit{Above})  Input images are passed through six localized attention blocks with shared weights to extract salient features. Three distance weighted localized attention steps are performed in each block. The features are forwarded to decoder blocks and MLP layer for binary prediction. (\textit{Below}) Inner working of localized attention modules. In our work we use $N=3$ blocks for each module.}
    \label{fig:network}
\end{figure}

The network architecture (Figure~\ref{fig:network}) utilizes localized attention modules, channel swaps, decoder blocks, and an multi-layer perceptron (MLP) layer to efficiently compare search and template images. At a high level, the model is organized into six stages of encoder-decoder pairs that ultimately feed into MLP aggregator for real-bogus prediction. The localized attention module extracts progressively higher-level features from the input image through successive applications of convolution and localized self-attention layers. By additionally exchanging channel information between search and template at the end of each encoding block, it allows downstream encoder blocks to learn salient spatial and channel features that help distinguish SNe from artifacts. The decoder block then fuses features from search and template localized attention modules and reduces the dimension for the prediction layer. Finally, the MLP collects all signals from six decoder blocks and concatenates them to make the real-bogus classification.

\subsection{Localized Attention Module}

The building block of the encoder network is the localized attention module that performs convolution, localized self-attention, and channel swap. It takes an input of dimension (N, $C_{in}$, $H_{in}$, $W_{in}$) and returns an output of dimension (N, $C_{out}$, $H_{out}$, $W_{out}$) where N is the batch size, C is the channel depth, H is the height, and W is the width.

To detect local features and allow for subsequent channel swap, we normalize the input through layer norm following the convolution step that increases the channel dimension on input $x_{in}$:
\begin{align*}
\mathbf{x}_{\mathrm{conv}}=\mathrm{LayerNorm}\!\left(\mathrm{Conv2D}(\mathbf{x}_{\mathrm{in}})\right)
\end{align*}

Only the first block reduces the spatial size by two pixels in each dimension. The remaining blocks keep $H_{\mathrm{in}}$ and $W_{\mathrm{in}}$ unchanged via zero-padding.

Following the convolution step, $x_{conv}$ is passed into three localized self-attention blocks that are the same as the original attention paper except for its re-weighting of the attention matrix by distance matrix D, with a learnable weight parameter $\alpha$ and the use of modified MLP \citep{NIPS2017_3f5ee243}. The attention element takes the form:
\begin{align*}
   \mathrm{Attention}(\mathbf{Q},\mathbf{K},\mathbf{V})
=\Big(\mathrm{Softmax}\!\big(\tfrac{\mathbf{Q}\mathbf{K}^{\top}}{\sqrt{d}}\big)\;\odot\;\exp(\alpha\,\mathbf{D})\Big)\,\mathbf{V},
\end{align*}

where $\mathbf{Q}$, $\mathbf{K}$, and $\mathbf{V}$ are query, key, and value, respectively.
The distance matrix $\mathbf{D}$ has dimensions ($N$, $1$, $H_{in}W_{in}$, $H_{in}W_{in}$) and contains pairwise pixel distances between pixels. The use of $\textbf{D}$ to localize attention in addition to the commonly used positional embedding guides the network to focus on extracting and seeking local feature comparisons between the two images. The addition of trainable parameter allows the network to learn the appropriate extent to localize the attention.

The modified MLP layer follows the same setup as CMT networks \citep{guo}. It inserts depth-wise convolution between the first feed-forward layer and GELU activation for more robust local feature extraction \citep{gelu}. Lastly, a channel swap is performed on the output of localized self-attention blocks $x_{attn}$ to mix the information from both images. The corresponding bottom half of the channels are exchanged between search and template images at every pixel location with this operation.
\begin{align*}
    [\textbf{y}_{i,src}, \textbf{y}_{i,tmp}] = \text{Swap}\left(\textbf{x}_{src,attn}, \textbf{x}_{tmp,attn}\right)
\end{align*}

\subsection{Decoder Block}

The decoder block collects intermediate signals from localized attention modules and outputs a signal for the MLP layer. The input signals from search and template images are first concatenated with a learnable position embedding. The search signal then undergoes a self-attention layer that uses the same modified MLP in the encoder blocks. We fuse the signals through cross-attention where search image is the query and a concatenated tensor of search and template acts as key and value. The fused signal undergoes another layer of self-attention and residual convolutional block. Lastly, a series of convolutional blocks followed by ReLU activation and batch normalization are applied.

\subsection{MLP}

The MLP layer concatenates six input signals from the decoder blocks and passes it through feed-forward neural network to make the final real-bogus prediction.

\subsection{Runtime}

During the training, the network processes $\sim$2 batches per second on an NVIDIA A100 GPU when the batch size is 64. This translates to about two hours of training per epoch for the entire \autoscan\ dataset. For inference, the memory requirements are significantly reduced and it can process a single batch of 1024 stamps in under 1.4 seconds. This is approximately the number of detections per CCD image for no-Diff dataset $\mathrm{SNR}$ cutoff and compares to $\sim$10 minutes per CCD processing time for \texttt{DiffImg} from DES \citep{diffimg}.

\section{Results}
\label{sec:results}

\subsection{\autoscan\ Dataset}

\begin{figure}[htbp]
    \centering
    \includegraphics[width=0.6\linewidth]{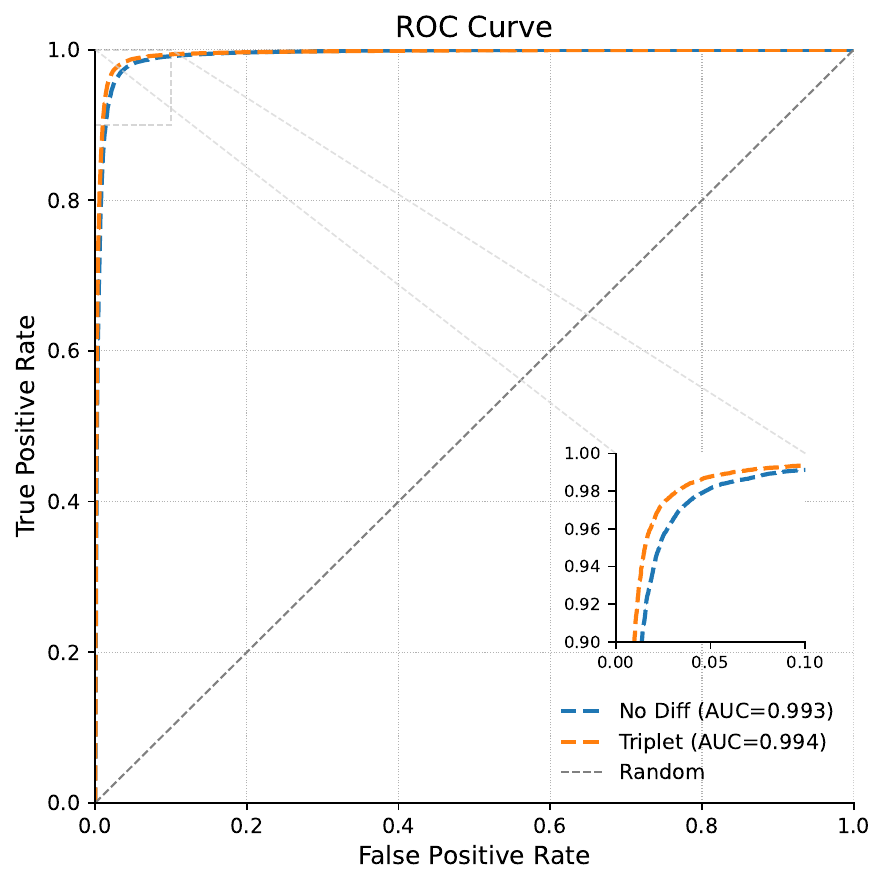}
    \caption{ROC curve for the \autoscan\ dataset with and without using difference image. The network trained on the entire dataset with a 9:1 train-test split.}
    \label{fig:roc_autoscan}
\end{figure}

\begin{figure}[htbp]
\gridline{\fig{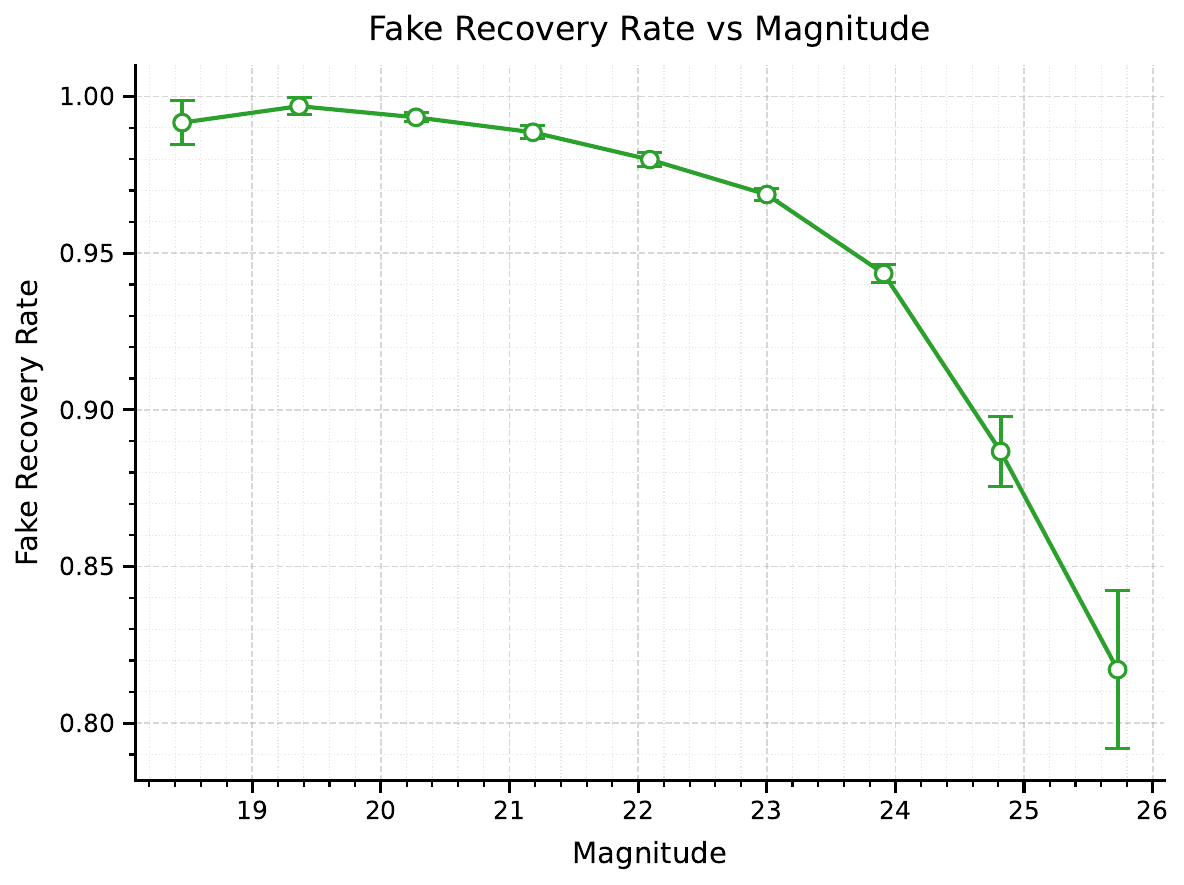}{0.49\textwidth}{}
          \fig{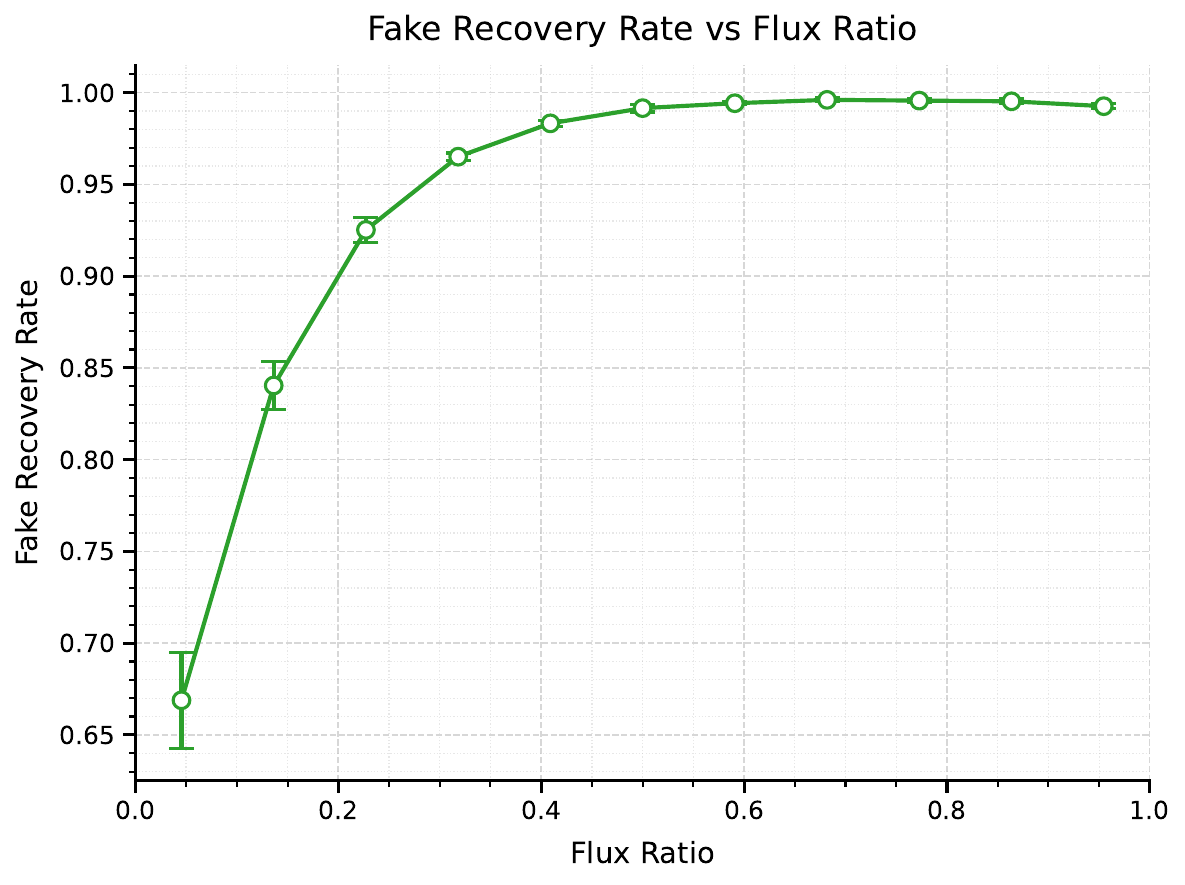}{0.49\textwidth}{}}
\caption{Fake recovery rate (true positive rate) analysis on the \autoscan\ test data. (\textit{Left}) Fake recovery rate as a function of magnitude. (\textit{Right}) Fake recovery rate as a function of flux ratio. The flux ratio is defined as $\text{clip}\left(0, 1, \frac{f_{d}}{f_t + f_{d}}\right)$ where $f_d$ and $f_t$ are fluxes of 5 by 5 central grids in difference and template images, respectively. Since both difference and template images are sky-subtracted, this ratio could fall outside the [0, 1] range. Bins with at least 25 test samples are shown in both panels.
\label{fig:autoscan_recovery}}
\end{figure}

We conducted two experiments using the \autoscan\ dataset to evaluate the performance of our neural network model. The first experiment focused on the network's ability to classify real and bogus supernova observations using only search and template images, bypassing the conventional reliance on difference imaging. Upon training the network with 800,768 samples across 50 epochs, we achieved an average classification accuracy of $97.4 \pm 0.2\%$ and ROC AUC of 0.993 (Figure~\ref{fig:roc_autoscan}) on five runs with test dataset of size 88,974. To further contextualize the network's performance across SN properties, we evaluated fake recovery (true-positive) rate across magnitude and flux ratio. We observe robust performance down to $\sim26$ mag (Figure~\ref{fig:autoscan_recovery}). However, these metrics likely understate the true performance, as a number of ``bogus'' labels in the test set are known to be incorrect \citep{acero-cuellar2023}. Upon manual inspection, a non-negligible fraction of images the network classified differently from the label could be reclassified. Adjusting for mislabeled data boosted the network's accuracy by approximately $0.2\%$.

The second experiment conducted a comparative analysis between two network configurations: one that utilized pairs of search and template images, and another that additionally incorporated the difference image. To ensure a fair comparison, the triplet network architecture was designed to closely mirror the search-template network, with minimal adjustments. Specifically, we used three parallel encoder blocks instead of two and adapted the channel swap mechanism to accommodate a three-way exchange.

\begin{figure}[htbp]
    \centering
    \includegraphics[width=0.7\linewidth]{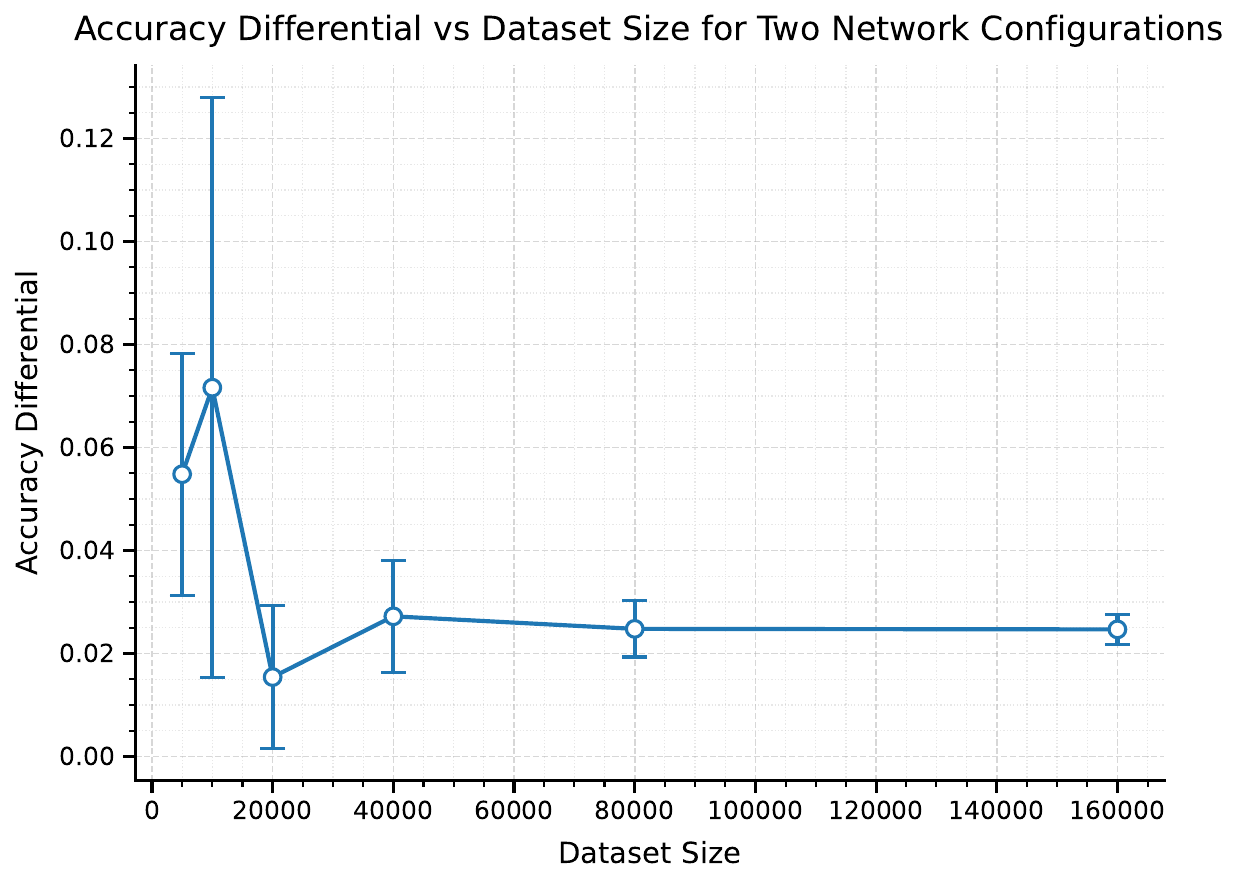}
    \caption{The difference in accuracy between search-template-diff network and search-template network as a function of dataset size for the \autoscan\ dataset after 10 epochs of training.}
    \label{fig:accuracy_diff}
\end{figure}

Results from this comparison reveal that the performance disparity between the two network setups narrows with increasing dataset size as shown in Figure~\ref{fig:accuracy_diff}. Specifically, in an experiment utilizing 800,768 search-template-difference triplets for training, the network configured with triplets attained a classification accuracy of 97.8\%—a modest improvement of 0.4 percentage points over its counterpart that processed only search-template pairs. This finding suggests that the benefits of including difference images may diminish as the size of the training dataset expands, potentially indicating that larger datasets can mitigate the absence of difference imaging.

We summarize our network performance and compare them to other published results in Table~\ref{tab:performance}. We include in the table a PyTorch implementation of \model{braai} network we trained to make a comparison to real-bogus classifier used at state of the art facility \citep{duev2019}. The \model{braai} network was trained under the same condition as our network.

\begin{deluxetable}{lcc}
  \label{tab:performance}
  \tablecaption{Performance comparison of networks trained on \autoscan\ dataset}
  \tablehead{
  \colhead{Network} & \colhead{Accuracy} & \colhead{ROC AUC}
  }
  \startdata
        Random Forest \citep{goldstein2015} & $96.8\%$ & - \\
        CNN without difference image \citep{acero-cuellar2023}\tablenotemark{a} & $91.1 \pm 0.5 \%$ & 0.973 \\
        CNN with difference image \citep{acero-cuellar2023}\tablenotemark{a} & $96.1 \pm 0.4 \%$ & 0.992 \\
        \model{braai} with difference image\tablenotemark{b} & $95.4 \pm 0.5 \%$ & 0.986 \\
        Attention without difference image (this work) & $97.4 \pm 0.2 \%$ & 0.993 \\
        Attention with difference image (this work) & $97.8 \pm 0.1 \%$ & 0.994 \\
  \enddata
  \tablenotetext{a}{Used only a subset of 100,000 samples rather than the full dataset.}
  \tablenotetext{b}{Trained a PyTorch implementation of \model{braai} under the same conditions as the Attention network.}
\end{deluxetable}

\FloatBarrier

\subsection{no-Diff Dataset}

\begin{figure}[htbp]
    \centering
    \includegraphics[width=0.5\linewidth]{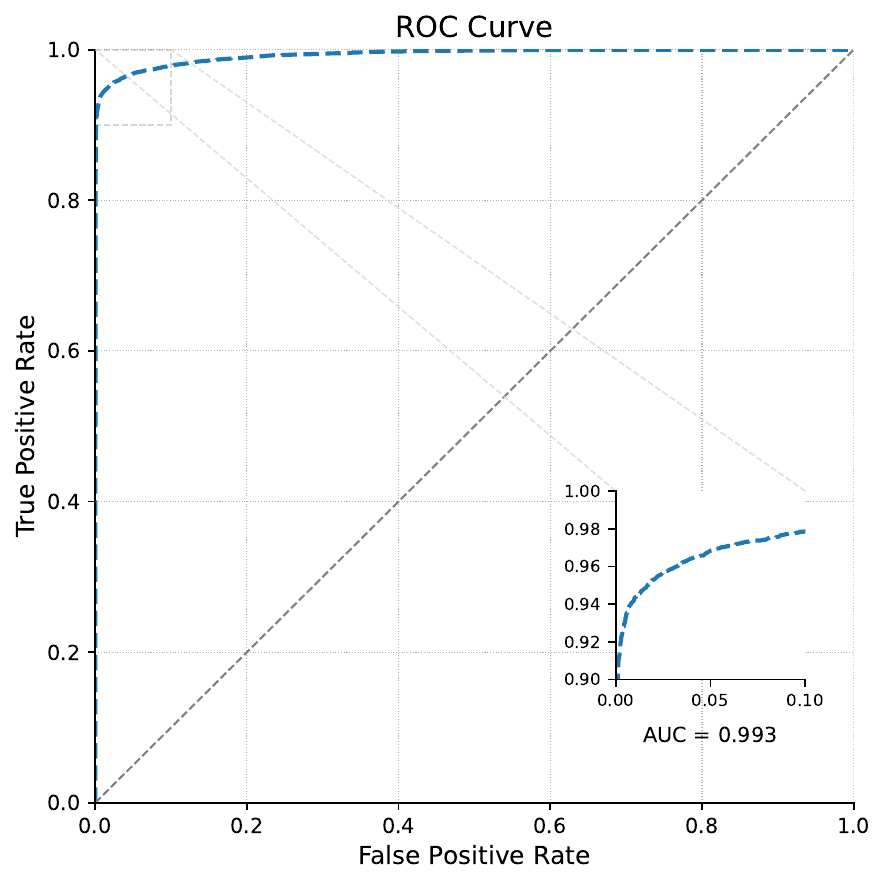}
    \caption{ROC Curve for no-Diff test dataset.}
    \label{fig:roc_nodiff}
\end{figure}

\begin{figure}[htbp]
\gridline{\fig{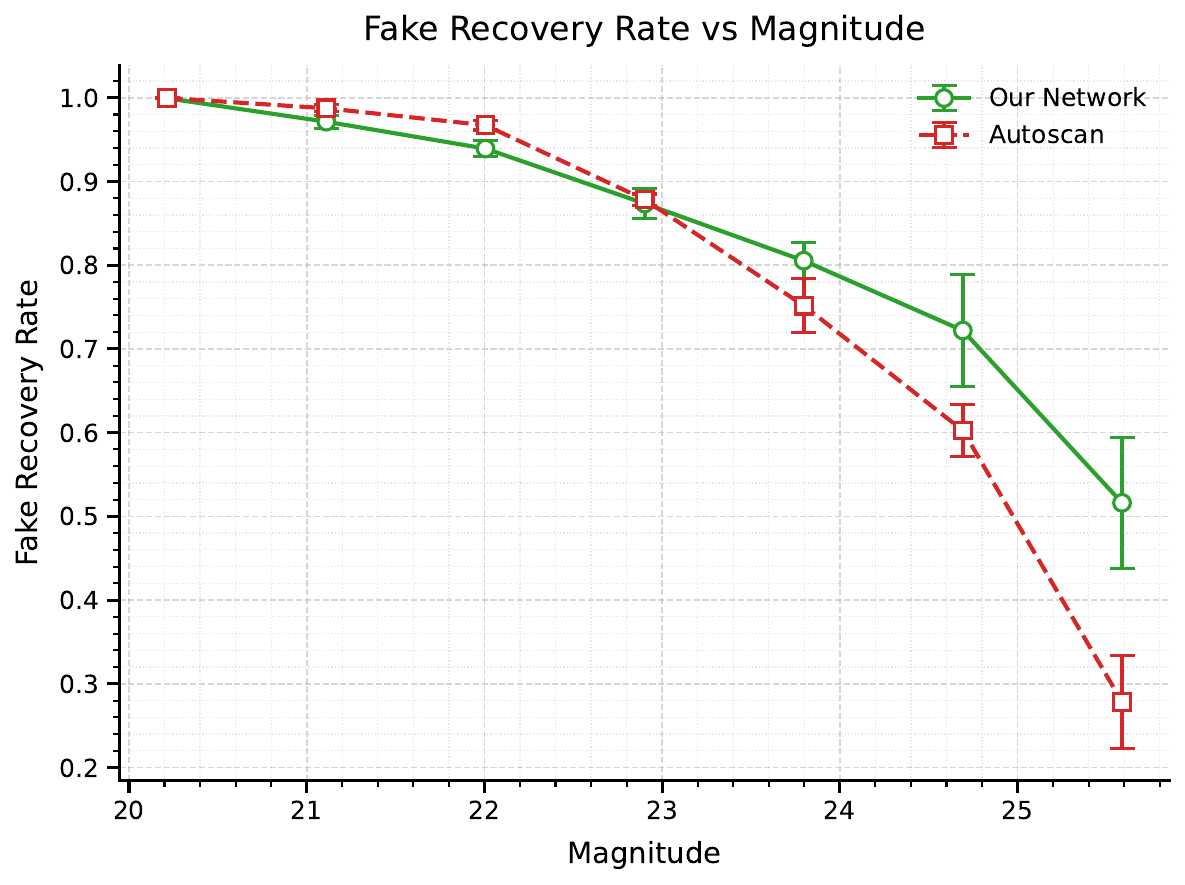}{0.45\textwidth}{}
          \fig{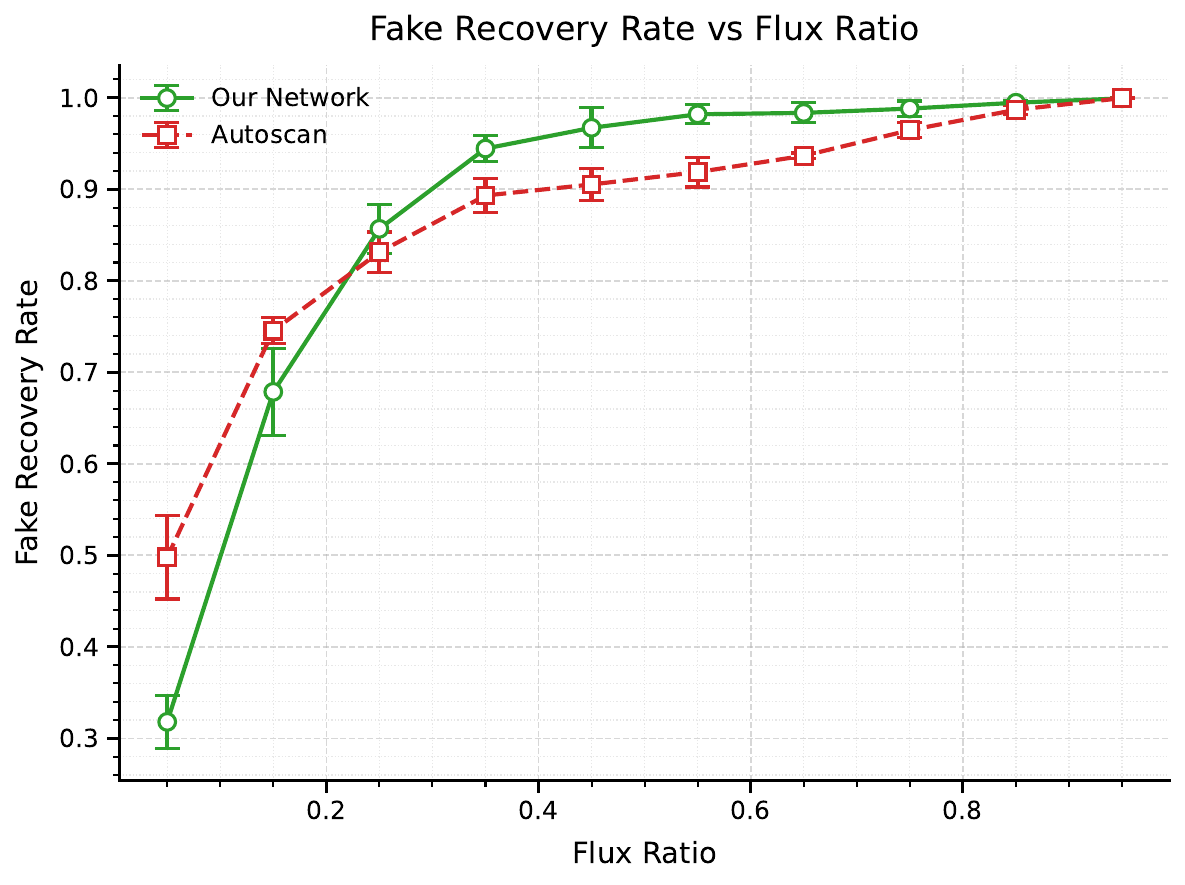}{0.45\textwidth}{}}
\caption{Fake recovery rate (true positive rate) analysis on the no-Diff dataset. (\textit{Left}) Fake recovery rate as a function of SN magnitude. Bins with at least 25 samples are shown. (\textit{Right}) Fake recovery rate as a function of flux ratio. The flux ratio is defined as $\text{clip}\left(0, 1, \frac{f_{d}}{f_t + f_{sn}}\right)$ where $f_{d}$ and $f_t$ are fluxes of 5 by 5 center grids centered on the SN in the difference and template images, respectively.
\label{fig:nodiff_recovery}}
\end{figure}

Our model achieved similar performance on the custom no-Diff dataset. The dataset consisting of 1,763,176 samples was first balanced to create a dataset of size 1,335,518. The balanced dataset was then further reduced to 20\% of its size and divided into a training and test set in a 9:1 ratio and trained for 50 epochs across five different seeds. The network achieved $96.7\pm0.1\%$ accuracy and an ROC AUC of $0.993\pm0.0$ (Figure~\ref{fig:roc_nodiff}). To bolster the robustness of the network, random horizontal and vertical flips of the images were performed during training. Fake recovery rate as a function of magnitude and flux ratio are shown in Figure~\ref{fig:nodiff_recovery}.

Although the results are not directly comparable with the \autoscan\ dataset given the differences in quality cuts, the available \autoscan\ scores allow us to make comparison within this dataset. Similar to the \autoscan\ dataset, the network outperforms \autoscan\ even when the source can appear anywhere in the image and only search and template images are provided. Across five seeds, fake recovery (true positive) rates for \autoscan\ and our network were $95.5\pm0.4\%$ and $95.8\pm0.5\%$. This gap widens when considering N-shot and majority consensus performance. For N-shot and majority consensus the recovery rates were $96.1 \pm 0.3\%$ and $96.1 \pm 0.3\%$, respectively.

\FloatBarrier

\section{Conclusions}
\label{sec:conclusions}

In this work, we introduced a transformer-based neural network for real-bogus classification that does not rely on difference imaging. We have demonstrated that a neural network can accurately distinguish real signals from artifacts without the need for difference image input and that the added value of a difference image diminishes with dataset size.

The study conducted on the \autoscan\ dataset highlights the effectiveness of the localized attention mechanism in capturing the salient changes between search and template images for correct real-bogus classification. The proposed method outperforms traditional approaches such as CNNs and random forests that rely on difference image information, and its performance approaches that of an attention network that leverages difference images. The latter lends support to the idea that the relevant information for accurate real-bogus classification resides in the search and template images alone.

We further investigated the effect of removing the implicit DIA information by creating the no-Diff dataset, where SNe are not guaranteed to be centered in the images. The network once again outperformed \autoscan\ and demonstrated its robustness to varying SN locations within the image.

Implementation of localized attention networks has the potential to bypass DIA in astronomical surveys and help scientists recover more signals from large-scale datasets. 

\section{Acknowledgement}

A.I. and M.S. were supported by DOE grant DE-FOA-0003177, NASA Grant NNH22ZDA001N-ROMAN, and NSF grant AST-2108094.

\bibliographystyle{apj}
\bibliography{references}

\end{document}